\documentclass[runningheads]{llncs}
\usepackage{graphicx}
\usepackage{comment}
\usepackage{float}
\usepackage{amsmath,amssymb} 
\usepackage{color}
\usepackage{multirow}
\usepackage{booktabs}
\usepackage{array}
\newcolumntype{C}[1]{>{\centering\arraybackslash}p{#1}}
\usepackage[pagebackref,breaklinks,colorlinks]{hyperref}
\usepackage{algorithm,algorithmicx}
\usepackage[noend]{algpseudocode}
\usepackage{algcompatible}
\usepackage{multirow}

\usepackage[skins]{tcolorbox} 
\tcbuselibrary{breakable} 
\newtcolorbox[auto counter, number within=section, list type=subsubsection, list inside=toc]{sectionbox}[2][]{
colback=white!98!gray, colframe=black, 
colbacktitle=white!90!gray, coltitle=black, 
fonttitle=\bfseries,
title={#2}, 
list entry={Comment \thetcbcounter\quad}
}

\usepackage{soul}

\definecolor{dt}{gray}{0.7}
\usepackage{makecell}

\makeatletter
\def\thickhline{%
	\noalign{\ifnum0=`}\fi\hrule \@height \thickarrayrulewidth \futurelet
	\reserved@a\@xthickhline}
\def\@xthickhline{\ifx\reserved@a\thickhline
	\vskip\doublerulesep
	\vskip-\thickarrayrulewidth
	\fi
	\ifnum0=`{\fi}}
\makeatother

\newlength{\thickarrayrulewidth}
\begin{document}
\pagestyle{headings}
\mainmatter
\renewcommand{\thefootnote}{}

\title{Incorporating Visual Experts to Resolve the Information Loss in Multimodal Large Language Models} 


\titlerunning{IVE: Incorporating Visual Experts}

\author{\!Xin He\textsuperscript{*},  Longhui Wei\textsuperscript{*$\dagger$},  Lingxi Xie, Qi Tian\textsuperscript{$\ddagger$}\!}

\authorrunning{X. He, L. Wei, et al.}

\institute{Huawei Inc.\\
\email{whut.hexin@gmail.com}, \email{weilh2568@gmail.com}, \email{198808xc@gmail.com},
\email{tian.qi1@huawei.com}}
\maketitle

\begin{abstract}
Multimodal Large Language Models (MLLMs) are experiencing rapid growth, yielding a plethora of noteworthy contributions in recent months. The prevailing trend involves adopting data-driven methodologies, wherein diverse instruction-following datasets are collected. However, a prevailing challenge persists in these approaches, specifically in relation to the limited visual perception ability, as CLIP-like encoders employed for extracting visual information from inputs. 
Though these encoders are pre-trained on billions of image-text pairs, they still grapple with the information loss dilemma, given that textual captions only partially capture the contents depicted in images.
To address this limitation, this paper proposes to improve the visual perception ability of MLLMs through a mixture-of-experts knowledge enhancement mechanism. Specifically, we introduce a novel method that incorporates multi-task encoders and visual tools into the existing MLLMs training and inference pipeline, aiming to provide a more comprehensive and accurate summarization of visual inputs. 
Extensive experiments have evaluated its effectiveness of advancing MLLMs, showcasing improved visual perception achieved through the integration of visual experts.

\keywords{Multimodal Large Language Models; Knowledge Enhancement; Integration of Visual Experts}

\end{abstract}

\footnote{* Equal contribution}
\footnote{$\dagger$ Project leader}
\footnote{$\ddagger$ Corresponding author}

\section{Introduction}

Recently, the development of large language models (LLMs)\cite{radford2018improving,radford2019language,touvron2023llama,touvron2023llama2} has notably propelled advancements in artificial general intelligence. Various domains within artificial intelligence have actively embraced LLMs to enhance their performance across different tasks\cite{ghosal2023text,liu2023visual,chen2023shikra,hong20233d}. The field of multimodal dialogue is no exception, witnessing a surge in the development of multimodal large language models (MLLMs) within recent months\cite{liu2023visual,zhu2023minigpt,wu2023visual,instructblip,2023visionllm,bai2023qwen,ye2023mplug,shen2023hugginggpt}. These works commonly insert visual extractors into LLMs, followed by fine-tuning a light-weight network to project extracted visual information into the language latent space.  

While recent advancements have notably elevated the performance of downstream multimodal dialogue tasks\cite{goyal2017making,singh2019towards,mishra2019ocr,masry2022chartqa,shah2019kvqa},
these improvements primarily stem from the collection of instruction data in various formats\cite{liu2023visual,liu2023mitigating,zhu2023minigpt,ye2023mplugdoc,chen2023shikra}. 
Pioneering works such as MiniGPT-4\cite{zhu2023minigpt} and LLaVA\cite{liu2023visual} introduced an automatic mechanism for generating general multimodal instruction data, leveraging the capabilities of ChatGPT\cite{chatgpt}. By subsequently fine-tuning MLLMs with the generated data, these approaches have achieved substantial enhancements in response quality for diverse queries.
Additionally, mPLUG-DocOwl\cite{ye2023mplugdoc} targets to amass instruction data related to documents, specifically enhancing the performance of MLLMs in document understanding tasks\cite{masry2022chartqa,mishra2019ocr,mathew2021docvqa}. Shikra\cite{chen2023shikra}, on the other hand, proposed to collect referring expression pairs and fine-tune MLLMs on these pairs, thereby strengthening the models' ability to handle the referential dialogue task. 
Furthermore, Instruct-BLIP\cite{instructblip} and other related works\cite{bai2023qwen,wang2023cogvlm} have proposed to assemble various multimodal datasets with distinct instruction templates. Subsequent fine-tuning of MLLMs on these consolidated datasets has proven instrumental in significantly improving their performances.

\begin{figure}[t]
    \centering
    \includegraphics[width=1.0\textwidth]{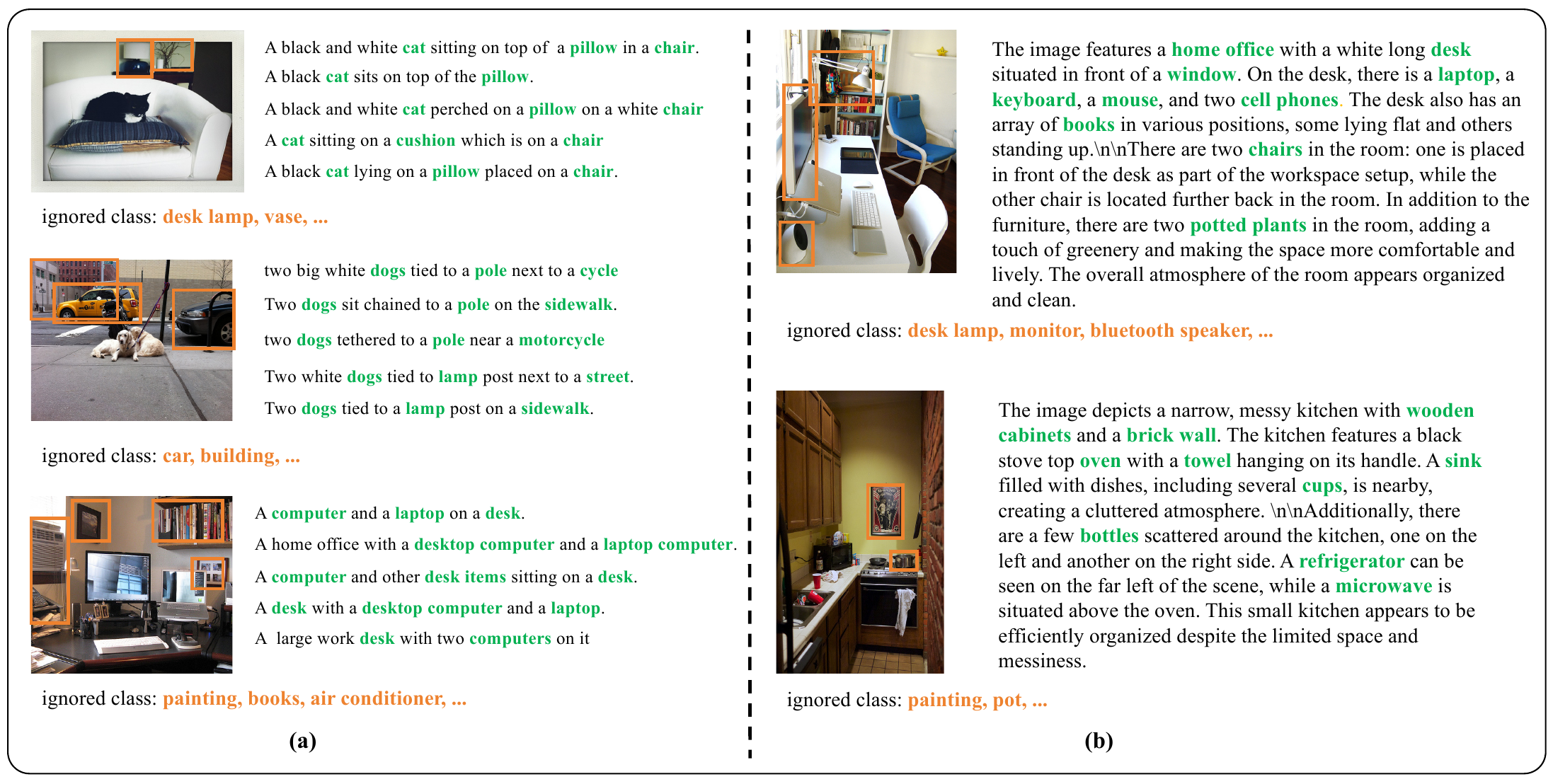}
    \caption{Examples from public image-text pairs.  (a) Examples from COCO Caption\cite{chen2015microsoft}. (b) Examples from 
LLaVA-Instruct-150K\cite{liu2023visual}.  The short textual captions in (a) make it difficult to comprehensively describe the corresponding image. The captions in (b) are more informative but still cannot describe the entirety of the image. The orange boxes in the image indicate objects that are ignored in the captions. }
    \label{fig:fig_1}
\end{figure}

As outlined above, while prior works have demonstrated advantages across various multimodal dialogue scenarios, they predominantly capitalize on the collected different types of instruction data, sharing a similar learning framework. Specifically, these works consistently employ a light-weight projection module (\emph{e.g.}, Q-Former in BLIP2\cite{li2023blip}) to map visual information, extracted by CLIP-like encoders (e.g., EVA-CLIP\cite{sun2023eva}), into the language latent space. Given that the CLIP-like encoders cannot comprehensively describe the entirety of visual inputs (for them pre-trained with short textual captions, as shown in Fig.\ref{fig:fig_1}(a)), MLLMs grapple with the information loss dilemma, which further restricts the response quality of queries. Moreover, though the detailed instruction data generated in LLaVA\cite{liu2023visual} or other works\cite{zhu2023minigpt,chen2023shikra}
can alleviate the above problem to some extent, there are still lots of details in images that cannot be fully described(as shown in Fig.\ref{fig:fig_1}(b)).
To address this challenge, there is a need for novel strategies that transcend the existing learning frameworks, enabling a more nuanced and accurate representation of visual information in MLLMs.

Inspired by the above, this paper explores MLLMs from the perspective of visual perception ability enhancement. Consequently, we introduce a simple but effective visual information learning framework, referred to as Incorporating Visual Experts (IVE), designed to augment the perception capabilities of MLLMs through aggregating available visual information extracted by specific experts.
Specifically, IVE mainly involves two additional modules, \emph{i.e.}, multi-task encoders and structural knowledge enhancement, for comprehensively describing the visual inputs.  The multi-task encoders integrate three auxiliary encoders, namely the low-level information encoder and the document-related information encoder, alongside with a CLIP-like encoder for semantics extraction. This integration aims to provide a more comprehensive description of visual inputs within the latent embedding space. The synergistic combination of these encoders facilitates a more nuanced understanding of the visual context. The structural knowledge enhancement mainly utilizes specific visual tools to extract structural data (e.g., the categories and locations of instances or textual information inside images). These structural data will serve as hard prompts and then be cooperated with the extracted latent embeddings fed into LLMs. More details about IVE have been presented in Sec.~\ref{sec:method}.

The introduced IVE is easy to implement, and its effectiveness has been substantiated through comprehensive experiments across various multimodal tasks. In general multimodal dialogue scenarios\cite{goyal2017making,marino2019ok}, IVE excels in recognizing the intrinsic content of input images, thereby producing more accurate responses to input queries in comparison to recent works. More results are expounded in Sec. \ref{sec:exper}. 
Furthermore, when applied to specific multimodal dialogue tasks such as DocVQA\cite{mathew2021docvqa}, IVE demonstrates competitive results when compared with recent state-of-the-arts. The above observations further demonstrate the improved visual perception achieved through the integration of visual experts.

\section{Related Work}
\subsection{Vision-and-Language Pre-training}
Most current multimodal large language models (MLLMs)\cite{liu2023visual,zhu2023minigpt,instructblip,ye2023mplug} are built on vision-and-language pre-training models (VLPs)\cite{li2020oscar,chen2020uniter,clip,li2022fine}, therefore we first revisit the development of VLPs before introducing MLLMs. The predominant VLP approaches can be broadly categorized into two frameworks: the one-stream framework\cite{li2020oscar,huang2020pixel,chen2020uniter,su2019vl} and the two-stream framework\cite{clip,mu2022slip,li2022fine,jia2021align}. Methods\cite{li2020oscar,huang2020pixel,chen2020uniter,su2019vl} within the one-stream framework typically employ a single transformer architecture to process both text and image data, incorporating various designs of loss functions. In contrast, the two-stream framework involves the independent extraction of modality information using distinct backbones. For instance, CLIP\cite{clip} utilizes a single image encoder for extracting visual information, while employing a textual encoder for processing textual information. For efficiency, current MLLMs\cite{zhu2023minigpt,instructblip,ye2023mplug} predominantly leverage the visual module of two-stream methods to encode the latent embeddings of visual inputs.

\subsection{Multimodal Large Language Models}

Multimodal Large Language Models (MLLMs) have garnered considerable attention from both academia and industries, with a surge in novel works emerging in recent months\cite{liu2023visual,instructblip,ye2023mplug}. A common framework underpins most of these works, featuring CLIP-like encoders responsible for extracting information from visual inputs, an abstractor summarizing the extracted information with few tokens, a light-weight layer further projecting the summarized information into the language latent space and a pre-trained large language model handling user questions in the context of the above extracted visual information. Despite their similar architectures, these works demonstrate versatility in addressing various multimodal dialogue tasks through training on distinct types of instruction data.
For instance, LLaVA\cite{liu2023visual} excels in generating detailed answers for generic images with training on comprehensive instruction data. On the other hand, mPLUG-DocOwl\cite{ye2023mplugdoc} achieves significant improvements in the performance of MLLMs on document analysis tasks by training on document-related instruction data. Shikra\cite{chen2023shikra} enhances the model's capability in handling referring questions by training on referring expression pairs. Although these works yield remarkable results, they remain constrained by the limited perception ability of CLIP-like encoders.
In contrast to previous approaches, this work takes a novel perspective by focusing on enhancing the visual perception ability of MLLMs. The proposed approach involves aggregating available visual experts to provide a more comprehensive description of visual inputs, aiming to overcome the constraints imposed by the existing limitations in visual perception ability.

\section{Our Approach}
\label{sec:method}
\subsection{Preliminaries}
Generally, the multimodal large language models (MLLMs)\cite{liu2023visual,instructblip,bai2023qwen,ye2023mplug} are usually composed of three modules, \emph{i.e.}, the visual perception module, the light-weight projection module, and the large language model, respectively. Specifically, the visual perception extracts the inside contents from visual inputs and then the light-weight projection module projects the above visual information into the language latent embedding space. The large language model module receives the projected visual information and generates textual responses for each query prompt. Therefore, given the visual inputs as ${x}_{i}$, the query as ${q}_{i}$, the visual perception module as $\mathrm{F_{vis}}(\cdot)$, the light-weight projection module as $\mathrm{F_{proj}}(\cdot)$ and the large language model as $\mathrm{LLM}(\cdot)$, the process of generating response in MLLMs can be formulated as:
\begin{eqnarray}
\label{eqn:problem_seeting}
\mathrm{Response}_{{q}_{i}:{x}_{i}} = \mathrm{LLM}(\mathrm{F_{proj}}(\mathrm{F_{vis}}({x}_{i})),{q}_{i}),
\end{eqnarray}
where $\mathrm{Response}_{{q}_{i}:{x}_{i}}$ denotes the generated response for the query ${q}_{i}$ based on the visual input ${x}_{i}$.

Restricted by the computing and data resources, most current MLLMs directly utilize well-trained large language models, such as Flan-T5\cite{chung2022scaling} and LLaMA\\\cite{touvron2023llama}, as the encyclopedia to answer the given question. Therefore, the key for MLLMs lies in how to properly summarize the information of visual inputs into language space. Currently, most MLLMs\cite{bai2023qwen,ye2023mplug,zhu2023minigpt} usually utilize CLIP-like encoders to extract the visual information, and then fine-tune a light-weight projection network with the collected instruction-following data to project extracted visual information into language latent space. Though extensive experiments have validated its effectiveness, the descriptions of visual inputs extracted by CLIP-like encoders are still not enough. As said "a picture is worth a thousand words", the CLIP-like encoders can only extract coarse semantic features inside each image in spite of their training on the billions of image-text pairs. 
To facilitate the above information loss dilemma, this paper proposes to incorporate visual experts in MLLMs, for comprehensively summarizing the visual contents of inputs. The details of our proposed approach will be carefully described in the next.

\subsection{Incorporating Visual Experts into MLLMs}
\begin{figure}
    \centering

    \includegraphics[width=1.0\textwidth]{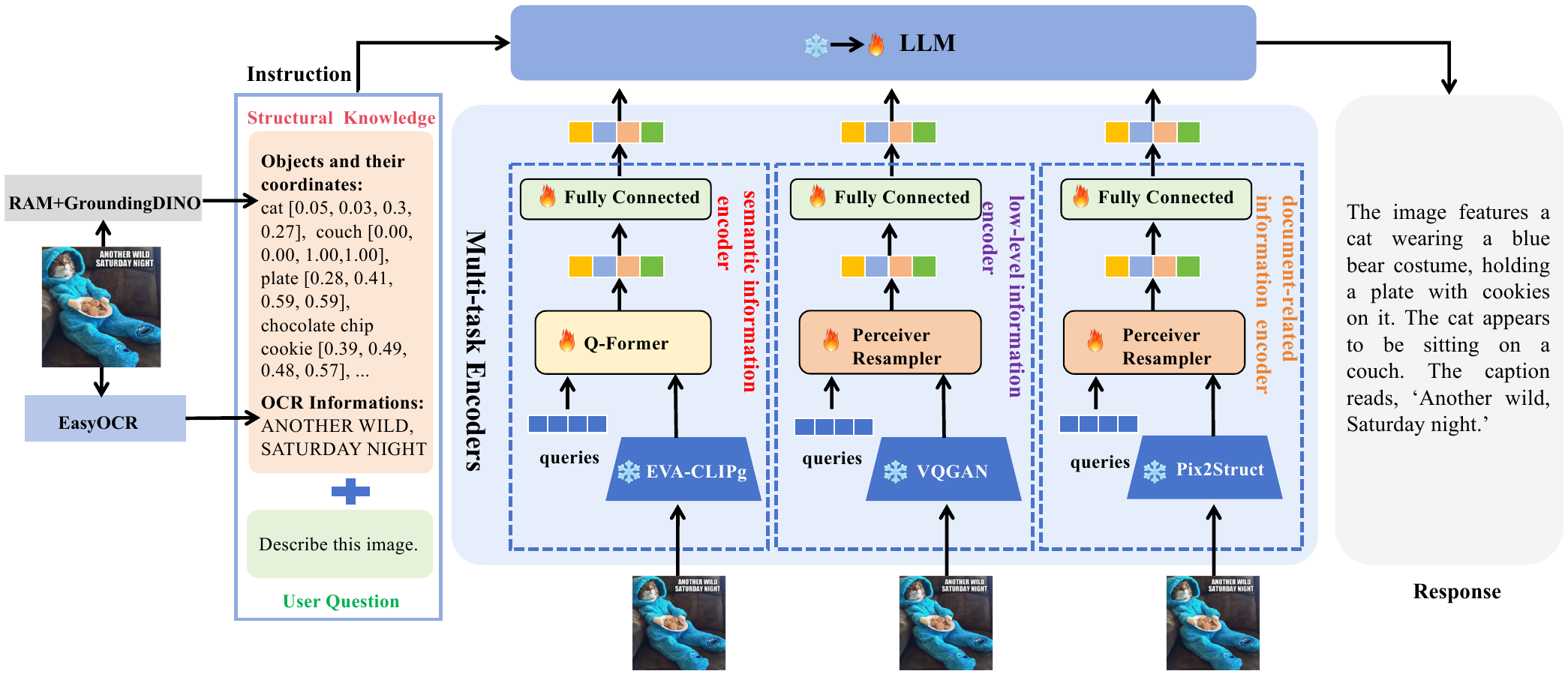}
    \caption{The illustrations of our proposed approach. Two modules, \emph{i.e.}, the multi-task encoders and structural knowledge enhancement, are specifically designed in our framework. The multi-task encoders integrate multiple types of complementary encoders to collaboratively capture the latent information within visual inputs, \emph{i.e.}, the semantic information encoder, the low-level information encoder and the document-related information encoder, respectively. In the structural knowledge enhancement module, our work mainly utilizes visual tools (RAM\cite{zhang2023recognize}+GroudingDINO\cite{liu2023grounding} and EasyOCR\cite{easyocr}) to detect the instances and textual information inside images as the prior knowledge fed into the large language model.}

    \label{fig:framework}

\end{figure}
Different from previous works, this paper improves the visual perception ability of MLLMs from the perceptive of knowledge enhancement, and thus proposes a simple but effective framework with primarily \textbf{I}ncorporating different types of \textbf{V}isual \textbf{E}xperts into the current MLLMs, referred as IVE. 
As shown in Fig. \ref{fig:framework}, the visual perception within IVE relies on two pivotal modules: multi-task encoders and structural knowledge enhancement module. 
The multi-task encoders are dedicated to amalgamating various types of latent visual information extracted by multiple visual encoders. This integration improves its comprehensiveness in the view of latent embedding. 
Additionally, the structural knowledge enhancement module is crafted to leverage visual tools, such as OCR tools\cite{easyocr} and object detectors\cite{zhang2023recognize,liu2023grounding}, to extract prior knowledge from visual inputs. This extracted knowledge is then treated as hard prompts and incorporated into the large language model alongside the previously fused latent embeddings.
Through the above cooperative modules, IVE can comprehensively encode the internal contents of visual inputs from diverse perspectives, thereby improving the quality of response to each query.

\subsubsection{Multi-task Encoders.}

The majority of current MLLMs commonly rely on CLIP-like encoders for extracting semantic information from visual inputs. However, the constrained perception ability associated with this approach limits their performance across various dialogue scenes. In contrast, IVE seeks to enhance this limitation by integrating multiple types of complementary encoders to collaboratively capture the latent information within visual inputs. As depicted in Fig. \ref{fig:framework}, three primary types of encoders are employed: the semantic information encoder, the low-level information encoder, and the document-related information encoder, each contributing distinct perspectives to the overall understanding of visual content.

The semantic information encoder is designed to extract the semantics from visual inputs and subsequently project them into the language embedding space. Consistent with prevalent methodologies\cite{chen2023shikra,2023visionllm,liu2023visual,li2023blip}, we adopt the CLIP-like encoder proposed in BLIP-2\cite{li2023blip}, where EVA-CLIPg\cite{sun2023eva} is initially employed to extract visual information, followed by the Q-former designed to condense this information into a concise representation using a few tokens.
Leveraging extensive training with abundant image-text pairs, this encoder generates embeddings adept at capturing the global semantic information of each visual input. The process of semantic feature extraction can thus be delineated as follows:
\begin{eqnarray}
\label{eqn:3_2_1}
F_s(x_i) = \mathrm{CrossAtt_{Q}}(\mathrm{Enc_{eva}}(x_{i}),\{T_0,T_1,...,T_m\}),
\end{eqnarray}
where $\mathrm{Enc_{eva}}$ denotes the visual encoder of EVA-CLIPg, $\mathrm{CrossAtt_{Q}}$ represents the operations in Q-Former, $\{T_0,T_1,...,T_m\}$ denotes the query tokens and $m$ is the sum of query tokens, respectively.

Given the brevity of captions that only provide a coarse description of the global semantics within each image, the semantic information extracted by Eq. (\ref{eqn:3_2_1}) is apparently insufficient. To enhance the richness of detailed information within the extracted latent embedding, a low-level information extractor is introduced as the supplement. In this paper, we adopt the encoder from VQGAN\cite{esser2021taming} as the corresponding low-level information extractor, which can encode images into latent embedding and then reconstruct them with the decoder of VQGAN. However, directly integrating the extracted embedding into MLLMs is costly because of its high dimensionality. Following Flamingo\cite{alayrac2022flamingo}, we also utilize several query tokens to summarize this latent embedding with Perceiver Resampler\cite{alayrac2022flamingo}, and the resultant tokens are then considered as low-level latent embedding. Consequently, the process of low-level information extraction can be formulated as:
\begin{eqnarray}
\label{eqn:3_2_2}
F_l(x_i) = \mathrm{CrossAtt_{PR}}(\mathrm{Enc_{vqgan}}(x_{i}),\{T_0,T_1,...,T_n\}),
\end{eqnarray}
where $\mathrm{Enc_{vqgan}}$ denotes the pre-trained encoder of VQGAN\cite{esser2021taming},  $\mathrm{CrossAtt_{PR}}$ represents the operations in Perceiver Resampler\cite{alayrac2022flamingo} and $n$ represents the sum of query tokens for low-level information, respectively.

While the aforementioned low-level information extractor contributes additional details upon the semantic embedding, it's noteworthy that both are trained on general images and may lack specificity for certain types, such as the document image. To address this, a document-related information encoder is incorporated into the latent embedding learning framework. In our framework, Pix2Struct\cite{lee2023pix2struct}, a recent state-of-the-art approach in document analysis tasks, is employed for this purpose. 
Similar to the low-level information encoder, several query tokens are employed to succinctly summarize the extracted document-related information using Perceiver Resampler\cite{alayrac2022flamingo}. Generally, the process of document-related information extraction can be formulated as:
\begin{eqnarray}
\label{eqn:3_2_3}
F_d(x_i) = \mathrm{CrossAtt_{PR}}(\mathrm{Enc_{pix}}(x_{i}),\{T_0,T_1,...,T_k\}),
\end{eqnarray}
where $\mathrm{Enc_{pix}}$ denotes the pre-trained encoder of Pix2Struct\cite{lee2023pix2struct} and $k$ represents the sum of query tokens for document-related information.

Consequently, the final fused latent embeddings of each image in IVE can be formulated:
\begin{eqnarray}
\label{eqn:problem_seeting}
{f}_{x_i}^l = [\mathrm{F_{proj}^{s}}(\mathrm{F_s}(x_i)); \mathrm{F_{proj}^{l}}(\mathrm{F_l}(x_i)); \mathrm{F_{proj}^{d}}(\mathrm{F_d}(x_i))],
\end{eqnarray}
where $\mathrm{F_{proj}^{s}}$, $\mathrm{F_{proj}^{l}}$ and $\mathrm{F_{proj}^{d}}$ represent the linear projection layer for semantic information extractor, low-level information extractor and document-related information extractor, respectively.

\subsubsection{Structural Knowledge Enhancement.}
In view of the fact that the query tokens for each extractor undergo training in an end-to-end fashion, ensuring that the summarized embeddings encompass the entirety of visual input remains a challenge. Thereby, this paper further introduces a structural knowledge enhancement module to explicitly extract structural data within each image using specific visual tools. Finally, these data are subsequently treated as prompts and fed into the large language model alongside the fused latent embeddings.

\begin{table*}[t]\centering
\setlength{\abovecaptionskip}{0.2cm}
\label{structural_knowledge_prompt}
\begin{minipage}{1.0\columnwidth}\vspace{0mm}
\centering
\begin{sectionbox}[]{Structural  Knowledge Template} 
    \centering
      \footnotesize
    \begin{tabular}{p{0.97\columnwidth} c}
In addition to the image content, it also provides possible objects contained in the image and their coordinates.\\

Objects and their coordinates:
\\
$(c_{0}, x_{0}^0, y_{0}^0, x_{0}^1, y_{0}^1),\; (c_{1}, x_{1}^0, y_{1}^0, x_{1}^1, y_{1}^1),\;...,\;(c_{q},x_{q}^0, y_{q}^0, x_{q}^1, y_{q}^1),$\\
There may be some OCR text information in the image.\\
OCR Information:\\
$t_{0},\;t_{1},\;...,\; t_{o},$\\
Please combine all above information when answering the question.
    \end{tabular}
\end{sectionbox}
\caption{The details of structural knowledge template. $(c_{i}, x_{i}^0, y_{i}^0, x_{i}^1, y_{i}^1)\;$represents the corresponding category and bounding boxes of the $i$-th instance detected by RAM\cite{zhang2023recognize}+GroundingDINO\cite{liu2023grounding}, $t_{i}\;$represents the $i$-th textual segment detected by EasyOCR\cite{easyocr}.
}
    \label{tab:struct_knowledge_prompts}
\end{minipage}
\end{table*}

Typically, human observation of an image involves first identifying the objects (their categories and locations) or textual information within this image. Drawing inspiration from this human cognitive process, the structured knowledge enhancement module is purposefully crafted to extract three types of information: the category and localization of instance, together with textual content, respectively.
We first utilize two specific visual tools (\emph{i.e.}, RAM\cite{zhang2023recognize} and Grounding DINO\cite{liu2023grounding}) to recognize and localize the objects inside each image.
Furthermore, we utilize EasyOCR\cite{easyocr} to detect the contained textual information of each visual input. Therefore, thanks to the above visual tools, most instances ${[(c_{0},x_{0}^0,y_{0}^0,x_{0}^1,y_{0}^1),...,(c_{q},x_{q}^0}$${,y_{q}^0,x_{q}^1,y_{q}^1)]}$ and textual information $[t_{0},t_{1},...,t_{o}]$ inside each image can be detected, where $c_{i}$ denotes the category of the detected $i$-th instance, $(x_{i}^0,y_{i}^0,x_{i}^1,y_{i}^1)$ represents the corresponding bounding boxes, $t_{i}$ means the detected $i$-th visual text segment, $q$ and $o$ are the sum of detected instances or textual segments, respectively. Thereby, the final extracted structural knowledge can be formulated as:
\begin{eqnarray}
\label{eqn:3_2_6}
\mathrm{f}_{x_i}^s  = [(c_{0},x_{0}^0,y_{0}^0,x_{0}^1,y_{0}^1),...,(c_{q},x_{q}^0,y_{q}^0,x_{q}^1,y_{q}^1);t_{0},t_{1},...,t_{o}],
\end{eqnarray}
To better align with LLM, we design the template in which inserting the extracted structural knowledge. The details of structural knowledge template have been shown in Tab.~\ref{tab:struct_knowledge_prompts}.

While extant literature, exemplified by LLaMA-Adapter v2\cite{gao2023llama}, has explored the integration of visual tools to extract structural knowledge with the aim of augmenting the visual perceptual capabilities of MLLMs, it is notable that these approaches\cite{gao2023llama,shen2023hugginggpt} have predominantly restricted the deployment of visual tools solely into the inference stage. In contrast, the proposed IVE is meticulously crafted to harness structural knowledge throughout both the training and inference phases of MLLMs. This strategic design of IVE serves the dual purpose of mitigating the inherent noise introduced by the visual tools and comprehensively capitalizing on the informative cues they provide.

\subsection{Training Pipeline 
\label{sec:3_3}
}
Once the fused latent embeddings and structural knowledge are available, we feed them into the large-scale language model (LLM) and conduct the overall training, which makes LLM better handle these prompts while ignoring the inevitable noises. Following previous works\cite{bai2023qwen,instructblip}, we reorganize the available public multimodal datasets\cite{marino2019ok,masry2022chartqa,mathew2021docvqa,liu2023visual}, and conduct supervised fine-tuning on them.
Overall, our model employs a three-stage training strategy: pretraining, multi-task instruct tuning, and specific fine-tuning. In the pretraining stage, we primarily utilize weakly labeled  image-text pairs to train the alignment module in the semantic information encoder. The multi-task instruct tuning stage involves training on various multimodal instruction datasets\cite{goyal2017making,marino2019ok,mishra2019ocr,mathew2021docvqa,shah2019kvqa}. Subsequently, in the specific fine-tuning stage, we fine-tune the model on selected specific datasets\cite{masry2022chartqa,mathew2021docvqa} to better adapt to their unique characteristics. Detailed descriptions of each training process are provided below.
\subsubsection{Stage 1: Pretraining.}
During this phase, we exclusively focus on training the Q-Former layer and its corresponding projection layer within the semantic information encoder. The low-level information encoder and document-related encoder are ignored in this stage. Moreover, the parameters of other modules remain frozen throughout this stage. Consistent with prevalent methodology\cite{li2023blip}, the input resolution for the semantic information encoder is set as 224$\times$224.
\subsubsection{Stage 2: Multi-task Instruct Tuning.}
Building upon Stage 1, we combine several public multimodal instruction datasets\cite{goyal2017making,marino2019ok,hudson2019gqa,shah2019kvqa,singh2019towards,mishra2019ocr,mathew2021docvqa,masry2022chartqa,krishna2017visual,liu2023visual,liu2023mitigating},  for multi-task instruct tuning. During this phase, we fine-tune the language model using LoRA\cite{hu2021lora}. The Q-Former, Perceiver Resampler, and their corresponding projection layers within the three encoders in our framework actively participate in training, while the parameters of other modules remain frozen. The input resolution for the semantic information encoder is increased to 448$\times$448, while the low-level information encoder is configured with the input resolution of 256$\times$256. Consistent with prevalent methodology\cite{lee2023pix2struct}, the input resolution of the document-related information encoder is set to 1024$\times$1024. The extracted structural knowledge is employed to enhance the comprehensiveness of visual inputs in this stage.
\subsubsection{Stage 3: Specific Fine-Tuning.}
In this stage, further fine-tuning is conducted on specific datasets to fit the unique characteristics of these datasets. Similar to the preceding stage, fine-tuning of the large language model (LLM) is executed using LoRA\cite{hu2021lora}. The Q-Former, Perceiver Resampler, and the corresponding projection layers in mutli-task encoders are trainable, while the parameters of all other modules remain frozen. Similar with Stage 2, the extracted structural knowledge is employed to enhance the comprehensiveness of visual inputs.

\section{Experiments}
\label{sec:exper}
\subsection{Datasets}
\subsubsection{Training Dataset.}
As mentioned in Sec.~\ref{sec:3_3}, the entire training pipeline comprises three stages. In Stage 1, about 300 million image-text pairs crawled from the Internet\cite{li2022fine} are initially utilized to train Q-Former (as the Stage 1 in BLIP-2). Subsequently, the LLaVA-CC3M-Pretrain-595K from LLaVA\cite{liu2023visual} is employed to further train Q-Former and the projection layer (as Stage 2 in BLIP-2). In Stage 2 of our framework, following previous work\cite{bai2023qwen}, multi-task datasets are incorporated, including several general VQA datasets (VQAv2\cite{goyal2017making}, OKVQA\cite{marino2019ok}, GQA\cite{hudson2019gqa}, KVQA\cite{shah2019kvqa}), OCR-related VQA datasets (TextVQA\cite{singh2019towards}, OCRVQA\cite{mishra2019ocr}), document-related VQA datasets (DocVQA\cite{mathew2021docvqa}, ChartQA\cite{masry2022chartqa}, WikiTableQuestions (WTQ)\cite{berant2019explaining}), grounding datasets (RefCOCO\cite{kazemzadeh2014referitgame}, RefCOCO+\\\cite{yu2016modeling}, RefCOCOg\cite{mao2016generation}, Visual Genome\cite{krishna2017visual}), image captioning datasets (COCO Caption\cite{chen2015microsoft}), and multimodal instruction datasets (LLaVA-Instruct-150K\cite{liu2023visual} and LRV-Instruction\cite{liu2023mitigating}). Additionally, Chinese-LLaVA-Vision-Instructions\cite{llava150kcn} and COCO-CN\cite{li2019coco} are also utilized to enhance the corresponding proficiency in Chinese, along with SynthDoG\cite{kim2022ocr} to improve the OCR capabilities. The statistics of the used training data in Stage 2 are presented in Appendix~\ref{app:data}. In Stage 3, further fine-tuning is conducted on specific datasets individually to fit the unique characteristics of them.

\subsubsection{Evaluation Dataset.}
Comprehensive assessments have been conducted to verify the performance of the proposed method across various tasks. The evaluations cover general object recognition, OCR recognition, chart and document recognition, as well as other multimodal dialogue tasks. The VQAv2\cite{goyal2017making} test set, OKVQA\cite{marino2019ok} test set, TextVQA\cite{singh2019towards} validation set, OCRVQA\cite{mishra2019ocr} test set, DocVQA\cite{mathew2021docvqa} validation set, ChartQA\cite{masry2022chartqa} test set, WTQ\cite{berant2019explaining} test set, and MME Benchmark\cite{fu2023mme} are chosen for the evaluations.
\subsection{Implementation Details}
\subsubsection{Model Configuration.}
Following previous work\cite{li2023blip}, the semantic information encoder in IVE adapts the EVA-CLIPg\cite{sun2023eva} as visual backbone, and the Q-Former is employed to distill this information into a concise representation using a limited number of tokens. In the low-level information encoder, we use  the encoder from VQGAN\cite{esser2021taming} to extract the low-level information. In the document-related information encoder, we use  the encoder from Pix2Struct-Large\cite{lee2023pix2struct} to extract document-related information. 
In the last two encoders, we respectively utilize a 3-layer and 6-layer Perceiver Resampler, both derived from Flamingo\cite{alayrac2022flamingo}, aimed at summarizing latent embeddings. Our multi-task encoders finally produce 128 visual tokens, with 32 tokens from the semantic information encoder, 32 tokens from the low-level information encoder, and 64 tokens from the document-related information encoder. Furthermore, these visual tokens undergo projection through linear project layers and input into LLaMA2-chat (7B)\cite{touvron2023llama2} for generating the corresponding responses.
\subsubsection{Training Details.}
IVE is structured around three training stages. In Stage 1, only the Q-Former and the projection layer of the semantic information encoder are trainable, while all other modules are held frozen. When training with the 300M image-text pairs\cite{li2022fine}, the training encompasses only 1 epoch, and a global batch size of $2048$. While training with the LLaVA-CC3M-Pretrain-595K\cite{liu2023visual}, the training encompasses 5 epochs, and a global batch size of $1024$. The learning rate in this stage employs a cosine warm-up strategy ($2000$ steps), with a maximum learning rate of $1e$-$4$,
and a minimum learning rate of $1e$-$6$. In Stage 2$\&$3, the language model undergoes fine-tuning using LoRA\cite{hu2021lora} with the parameters of rank=$64$. The Q-Former, Perceiver Resampler, and their corresponding projection layers are involved in training, while the parameters of other modules remain frozen. In the last two stages, the training encompasses $1$ epoch, and a global batch size of $128$. As for the learning rate, we employ a cosine warm-up strategy ($500$ steps), with a minimum learning rate of $1e$-6 and a maximum learning rate of $3e$-5 for Stage 2, $1e$-5 for Stage 3. AdamW\cite{loshchilov2017decoupled} serves as the optimizer for all three training stages, with $\beta1=0.9$, $\beta2=0.98$, and the weight decay of $0.05$.
\subsection{Direct-transfer performance on Visual Question Answer}
The VQA task entails the model answering questions based on both the input image and query. In this section, we conduct direct-transfer evaluations on multiple VQA benchmarks using the IVE model trained after multi-task instruct tuning stage. We compare the proposed methods with several state-of-the-arts, including Qwen-VL-Chat\cite{bai2023qwen}, mPLUG-DocOwl\cite{ye2023mplugdoc}, mPLUG-Owl2\cite{ye2023mplug2}, and LLaVA-1.5\cite{liu2023improved}. The evaluation encompasses seven benchmarks: VQAv2\cite{goyal2017making} and OKVQA\cite{marino2019ok} for the general VQA task, TextVQA\cite{singh2019towards} and OCRVQA\cite{mishra2019ocr} for the OCR VQA task, and ChartQA\cite{masry2022chartqa}, DocVQA\cite{mathew2021docvqa}, and WTQ\cite{berant2019explaining} for the document or chart VQA task. Consistent with Stage 2\&3 in training phase, we employ the following prompt for all VQA evaluations: “\textless Img\textgreater \{latent embedding\}\textless/Img\textgreater \\ \{structural knowledge\}\{question\}. Answer the question using a single word or phrase." In addition, as the object detection results of the chart and document images are usually useless, we design an automatic filtering mechanism to filter out the detection results of these images. 
\begin{table}[t]
\centering
\caption{The direct-transfer results on VQA datasets.}
\scalebox{0.72}{
\begin{tabular}{lc|ccccccc}
\toprule[0.4mm]
 Model & LLM & VQAv2\cite{goyal2017making} & OKVQA\cite{marino2019ok} & TextVQA\cite{singh2019towards} & ChartQA\cite{masry2022chartqa} & OCRVQA\cite{mishra2019ocr} & WTQ\cite{berant2019explaining} & DocVQA\cite{mathew2021docvqa} \\ \midrule[0.4mm]
 
 BLIP-2 \cite{li2023blip} & 13B & 65.0 &  45.9 & 42.4 & - &  -  & - & - \\
 InstructBLIP \cite{instructblip} & 13B & - &  - & 50.7 & - &  -  & - & - \\
 Shikra \cite{chen2023shikra} & 13B & 77.4 &   47.2 & - & - &  -  & - & - \\
 mPLUG-DocOwl\cite{ye2023mplugdoc} & 7B & - & - & 52.6 & 57.4 & - & 26.9 & 62.2 \\
  Qwen-VL-Chat\cite{bai2023qwen} & 7B & 78.2 & 56.6 & 61.5 & \textbf{66.3} &  70.5  & - & 62.6 \\
  LLaVA-1.5\cite{liu2023improved} & 7B & 78.5 & - & 58.2 & - & - & - & -\\
  mPLUG-Owl2\cite{ye2023mplug2} & 7B & \textbf{79.4} & 57.7 & 58.2 & - & - & - & -  \\
  
  \textbf{IVE(ours)} & 7B & 78.8 & \textbf{60.3} & \textbf{62.0} & 65.3& \textbf{71.1} & \textbf{29.8} & \textbf{64.1} \\
  
  \bottomrule[0.4mm]
\end{tabular}
}
\label{tab:direct-transfer}
\end{table}
\indent 

As indicated in Tab.~\ref{tab:direct-transfer}, our model demonstrates competitive performance when compared to recent approaches. Specifically, IVE achieves an accuracy of $60.3\%$ on OKVQA, which significantly surpasses the performance of recent state-of-the-art method (mPLUG-Owl2\cite{ye2023mplug2} achieved $57.7\%$). In TextVQA\cite{singh2019towards} and OCRVQA\cite{mishra2019ocr} datasets, IVE achieves accuracies of $62.0\%$ and $71.1\%$, outperforming Qwen-VL-Chat\cite{bai2023qwen} with $0.5\%$ and $0.6\%$, respectively. As for the DocVQA\cite{mathew2021docvqa}, and WTQ\cite{berant2019explaining} datasets, IVE still achieves consistent improvements compared with recent approaches. More visualized examples have been shown in Appendix~\ref{app:vis}.

\subsection{Fine-tuning on Visual Question Answer}
To compare our model with specific VQA methods, we assess the performance of IVE further fine-tuning on the VQAv2\cite{goyal2017making}, OKVQA\cite{marino2019ok}, OCRVQA\cite{mishra2019ocr}, and ChartQA\cite{masry2022chartqa}. We still employ the prompt: “\textless Img\textgreater \{latent embedding\}\textless/Img\textgreater\\ \{structured knowledge\}\{question\}. Answer the question using a single word or phrase."  during evaluation. The further fine-tuning results of IVE on these VQA datasets are shown in Tab.~\ref{tab:fine-tuning}.
\begin{table}[t]

\centering
\caption{The fine-tuning results on VQA datasets.}
\scalebox{0.85}{
\begin{tabular}{lc|cccc}
\toprule[0.4mm]
 Model & LLM & VQAv2\cite{goyal2017making} & OKVQA\cite{marino2019ok} & OCRVQA\cite{mishra2019ocr} & ChartQA\cite{masry2022chartqa} \\ \midrule[0.4mm]
 BLIP2\cite{li2023blip} & 13B  & 82.2 & 59.3 & 72.7 & -  \\
 GIT\cite{wang2022git} & - & 78.6 & - & 68.1 & -  \\
 GIT2\cite{wang2022git} & - &  81.7 & - & 70.3 & -\\
 InstructBLIP\cite{instructblip} & 13B & - & 62.1 & 73.3 & -  \\
  CogVLM\cite{wang2023cogvlm} & 7B & \textbf{84.7} & 64.7 & 74.5 & - \\
  Pix2Struct-Large\cite{lee2023pix2struct} & - & - & - & 71.3 & 58.6   \\
  
  \textbf{IVE(ours)} & 7B & 84.0 & \textbf{65.2} & \textbf{74.9} & \textbf{68.3}  \\
  
  \bottomrule[0.4mm]
\end{tabular}
}
\label{tab:fine-tuning}
\end{table}

The experimental results demonstrate that our method, following additional fine-tuning on specific datasets, achieves favorable outcomes. Specifically, there are $5.2\%$ and $4.9\%$ improvements compared with the direct-transfer results on VQAv2\cite{goyal2017making} and OKVQA\cite{marino2019ok}. 
Notably, in tasks related to OCR and charts, IVE significantly outperforms the Pix2Struct\cite{lee2023pix2struct} method in OCRVQA\cite{mishra2019ocr} and ChartVQA\cite{masry2022chartqa}, with $3.6\%$ and $9.7\%$ improvements, respectively. Additionally, when compared to the recent state-of-the-art (CogVLM\cite{wang2023cogvlm}), IVE still shows competitive results.

Given that the MME Benchmark\cite{fu2023mme} focuses on the yes/no QA format, we conduct further fine-tuning of our multi-task instruct tuning model using a mixed dataset composed of VQAv2\cite{goyal2017making} and LRV-Instruction \cite{liu2023mitigating}. Subsequently, we evaluate the model on the MME Benchmark. As demonstrated in Tab.~\ref{tab:mme}, our method achieves the scores of 1455.6 and 384.1 in the perception and cognition task of MME Benchmark\cite{fu2023mme}, respectively. Compared with recent state-of-the-arts (mPLUG-Owl2\cite{ye2023mplug2} and LLaVA-1.5\cite{liu2023improved}), our IVE demonstrates superior stability across these two tasks.

\begin{table}[t]

\centering
\caption{The evaluations on MME Benchmark.}
\scalebox{0.90}{
\begin{tabular}{lc|cc}
\toprule[0.4mm]
 Model & LLM & Perception & Cognition  \\ \midrule[0.4mm]
 mPLUG-Owl\cite{ye2023mplug} & 7B & 967.3 & 276.1   \\
 LRV-Instruction\cite{liu2023mitigating} & 7B  & 1299.8 & 328.2   \\
  Qwen-VL-Chat\cite{bai2023qwen} & 7B & 1487.6 & 360.7   \\
  LLaVA-1.5 \cite{liu2023improved} & 7B & \textbf{1510.7} & -  \\
  mPLUG-Owl2\cite{ye2023mplug2} & 7B & 1450.2 & 313.2    \\
  
  \textbf{IVE(Ours)} & 7B & 1455.6 & \textbf{384.1}   \\
  
  \bottomrule[0.4mm]
\end{tabular}
}
\label{tab:mme}
\end{table}

\subsection{Ablation Study}
To better evaluate the effectiveness of the proposed multi-task encoders and structural knowledge enhancement in IVE, we further conduct ablation studies with the experiments using 50\% of the mixed dataset in Stage 2 for efficiency.

\begin{table}[b]
\centering
\caption{The ablation studies of each proposed module on VQA datasets.}
\scalebox{0.76}{
\begin{tabular}{l|cccc}
\toprule[0.4mm]
 Methods & VQAv2\cite{goyal2017making} & TextVQA\cite{singh2019towards} & DocVQA\cite{mathew2021docvqa} & MME\cite{fu2023mme} \\ \midrule[0.4mm]
 semantic information encoder only  & 67.1 & 43.8 & 39.3 & 1145.6/276.7  \\
  + low-level information encoder  & 67.7 & 44.0 & 40.2 & 1180.3/292.3  \\
  + document-related information encoder & 68.2 & 46.3 & 43.6 & 1232.6/317.0 \\
  + structural knowledge enhancement on Infer& 67.9 & 47.4 & 43.3 & 1201.6/323.6   \\
  
  + structural knowledge enhancement on Train\&Infer & \textbf{70.6} & \textbf{50.8} & \textbf{45.1} & \textbf{1273.6}/\textbf{337.1}  \\
  
  \bottomrule[0.4mm]
\end{tabular}
}
\label{tab:ab}
\end{table}
\subsubsection{Effectiveness of Multi-task Encoders.}

To evaluate the individual contributions of each encoder within our multi-task encoders, three distinct experiments have been conducted. The initial experiment exclusively employs the semantic information encoder. Subsequently, in another experiment, both the semantic information encoder and low-level information encoder have been concurrently utilized. The final experiment involves the simultaneous deployment of the semantic information encoder, the low-level information encoder, and the document-related information encoder, thereby examining the combined effects of these components.

The experimental results in Tab.~\ref{tab:ab} demonstrate that fusing the semantic information encoder and  the low-level information encoder leads to improvements across various datasets compared to only using the semantic information encoder. Further fusion with the document-related information encoder results in a significant improvement on OCR and document VQA datasets, with TextVQA\cite{singh2019towards} rising from $44.0\%$ to $46.3\%$ and DocVQA\cite{mathew2021docvqa} rising from $40.2\%$ to $43.6\%$, respectively. More qualitative results have been present in Appendix~\ref{app:vis}.

\subsubsection{Effectiveness of Structural Knowledge Enhancement.}
To validate the effect of structural knowledge enhancement and compare the different impacts of integrating structural knowledge only in the inference phase or in both the training and inference phases, we further conduct two additional experiments built upon the multi-task encoders.

As shown in Tab.~\ref{tab:ab}, the performance on certain datasets, such as VQAv2\cite{goyal2017making} and DocVQA\cite{mathew2021docvqa}, experiences a degradation when incorporating structural knowledge solely during the inference phase. Conversely, integrating this expert knowledge during both the training and inference phases yields improved results across a spectrum of datasets.
The aforementioned outcomes suggest that the supplementary knowledge introduces inherent noises, negatively impacting response quality while it is directly utilized. However, when introducing these extracted knowledge during the training phase, the LLM is guided to autonomously discern and extract pertinent information, thereby mitigating the adverse effects of noise. Further qualitative results are presented in Fig.~\ref{fig:ab_figure}(b) and  Fig.~\ref{fig:ab_figure}(c) of Appendix~\ref{app:vis}.

Moreover, to demonstrate that integrating structural knowledge during both training and inference phases can mitigate the disturbation of noises in knowledge rather than simply aligning prompt formats, we conduct 
additional experiments with fine-tuning on the sampled VQAv2\cite{goyal2017making} dataset. 
Specifically, we replace the automatically detected results with the ground truth as our structural knowledge. We compare the result of integrating structural knowledge only in the inference phase or in both training and inference phases.
\begin{table}[t]
\centering
\caption{The ablation studies while regarding ground truth as the utilized structural knowledge.}
\scalebox{0.90}{
\begin{tabular}{l|c}
\toprule[0.4mm]
 Model & VQAv2\cite{goyal2017making} \\ \midrule[0.4mm]
 Multi-task Encoders  & 75.2  \\
  + structural knowledge enhancement on Infer  & 77.5   \\
  + structural knowledge enhancement on Train\&Infer & 77.9  \\
  
  \bottomrule[0.4mm]
\end{tabular}
}
\label{tab:ablation_GT}
\end{table}
As shown in Tab.~\ref{tab:ablation_GT}, utilizing the ground truth as structural knowledge and integrating it during both training and inference phases only achieves slight gains (0.4\%) compared to the mechanism of integrating ground truth during the inference phase. This observation suggests that our proposed method goes beyond simple prompt format alignment. Instead, it focuses on autonomously discerning and extracting pertinent information, thereby mitigating the adverse effects of noise.

\section{Conclusion}
This paper firstly reevaluates the existing limitations within current multimodal large language models(MLLMs), and points out that they always grapple with the information loss dilemma. To enhance the corresponding visual perception ability of MLLMs, we present Incorporating Visual Experts(IVE), the first work to aggregate available visual information through a mixture-of-experts mechanism in both training and inference stages. Extensive experiments on a wide range of multimodal dialogue datasets have evaluated the effectiveness of IVE. In the future, the unified interactive multimodal large language model with more visual experts enhancements will be explored.
\bibliographystyle{splncs04}
\bibliography{eccv}
\newpage

\appendix
\section{Multi-task Instruct Tuning Data.}
\label{app:data}
The details of our utilized instruction datasets in Stage 2 are presented in Tab.~\ref{tab:instruct_tuning_data}. Various multimodal datasets are collected to train IVE for enhancing its generalization on different multimodal dialogue scenarios.
\begin{table}[h]
    \centering
    \caption{Summary of multi-task instruct tuning data. 
    }
    \scalebox{0.70}{
    \begin{tabular}{l c l}
         \toprule[0.4mm]
         \textbf{Task} & \textbf{\# Samples} & \textbf{Dataset} \\
         \midrule[0.4mm]
             
         VQA            & 1.70M  & \makecell[l]{VQAv2\cite{goyal2017making}, OKVQA\cite{marino2019ok}, GQA\cite{hudson2019gqa}, KVQA\cite{shah2019kvqa}, TextVQA\cite{singh2019towards}, OCRVQA\cite{mishra2019ocr}, \\DocVQA\cite{mathew2021docvqa}, WTQ\cite{berant2019explaining}, ChartQA\cite{masry2022chartqa}}\\ 
         Captioning        & 0.40M  & 
         COCO Captioning\cite{chen2015microsoft}, COCO-CN\cite{li2019coco}\\ 
         Grounding      & 3.89M  & 
         RefCOCO\cite{kazemzadeh2014referitgame}, RefCOCO+\cite{yu2016modeling}, RefCOCOg\cite{krishna2017visual}, Visual Genome\cite{krishna2017visual}\\
         OCR            & 1.00M  & 
         SynthDoG-en\cite{kim2022ocr}, SynthDoG-zh\cite{kim2022ocr}\\
         Conversation            & 0.65M  &
         LLaVA-Instruct-150K\cite{liu2023visual},  LRV-Instruction\cite{liu2023mitigating}, 
         Chinese-LLaVA-Vision-Instructions\cite{llava150kcn}\\
          
         \bottomrule[0.4mm]
    \end{tabular}
    }
    \label{tab:instruct_tuning_data}
\end{table}

\section{Ablation Studies on Structural Knowledge Enhancement}
\label{app:ab}
\begin{figure}[t]
    
    \centering

    \includegraphics[width=1.0\textwidth]{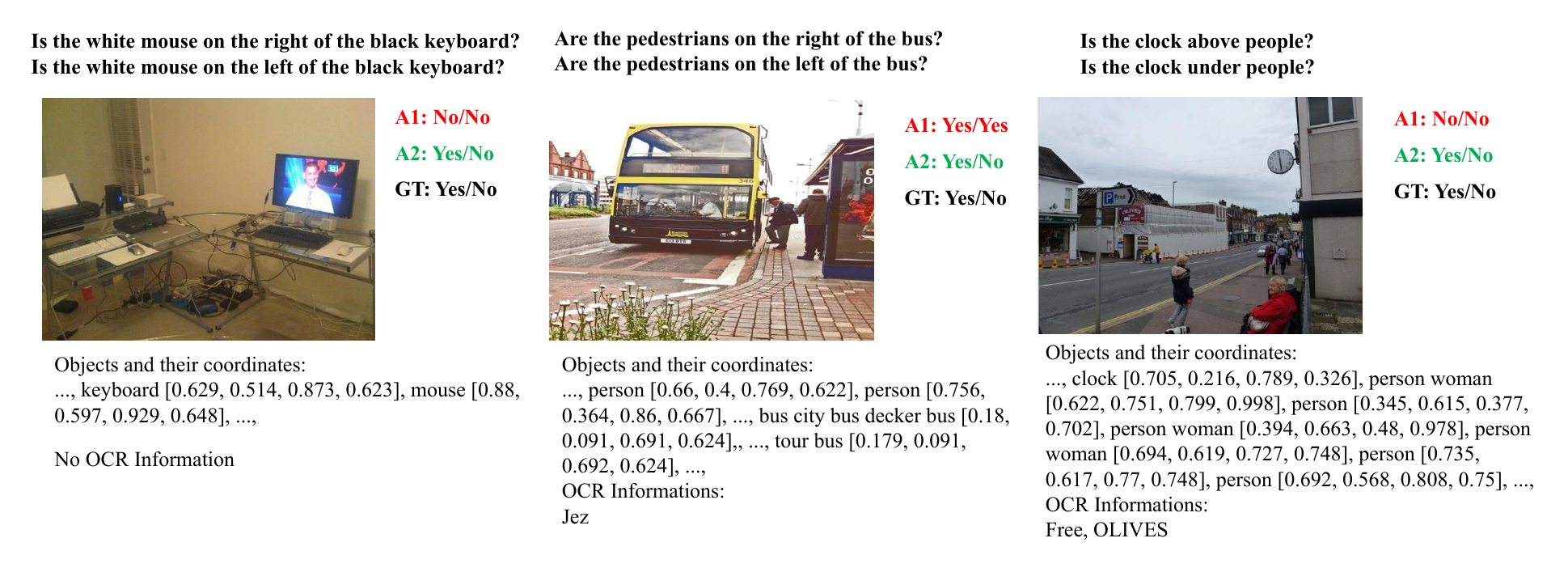}
    \caption{The qualitative analysis of structural knowledge enhancement on improving spatial awareness ability. A1 represents the result while not integrating structural knowledge, A2 represents the result while integrating structural knowledge in both training and inference stages, and GT represents the ground truth, respectively. The red lines represent the wrong answers and the green lines denote the correct answers.}
    \label{fig:ab_mme_pos}

\end{figure}

The structural knowledge extracted by Grounding DINO\cite{liu2023grounding} includes the coordinates of each detected instance, further representing their spatial relationships. To assess the effectiveness of this structural knowledge in enhancing spatial awareness capabilities, we conduct additional experiments on the MME Benchmark\cite{fu2023mme}. 
As shown in Tab.~\ref{tab:mme_pos_eva}, integrating structural knowledge during both training and inference phases can improve the accuracy from $75.0\%$ to $85.5\%$ on the position perception task in MME.
More visualized results have been shown in Fig.~\ref{fig:ab_mme_pos}.
\begin{table}[h]
\centering
\caption{The ablation studies of each proposed module on the position perception task in MME.}
\scalebox{0.90}{
\begin{tabular}{l|c}
\toprule[0.4mm]
 Model & MME(Position) \\ \midrule[0.4mm]
 Multi-task encoders  & 75.0    \\
    +structural knowledge enhancement on Infer   & 71.3   \\
  +structural knowledge enhancement on Train\&Infer & 85.5  \\
  \bottomrule[0.4mm]
\end{tabular}
}
\label{tab:mme_pos_eva}
\end{table}

\section{Visualizations.}
\label{app:vis}

\begin{figure}
    \centering
    \includegraphics[width=1.0\textwidth]{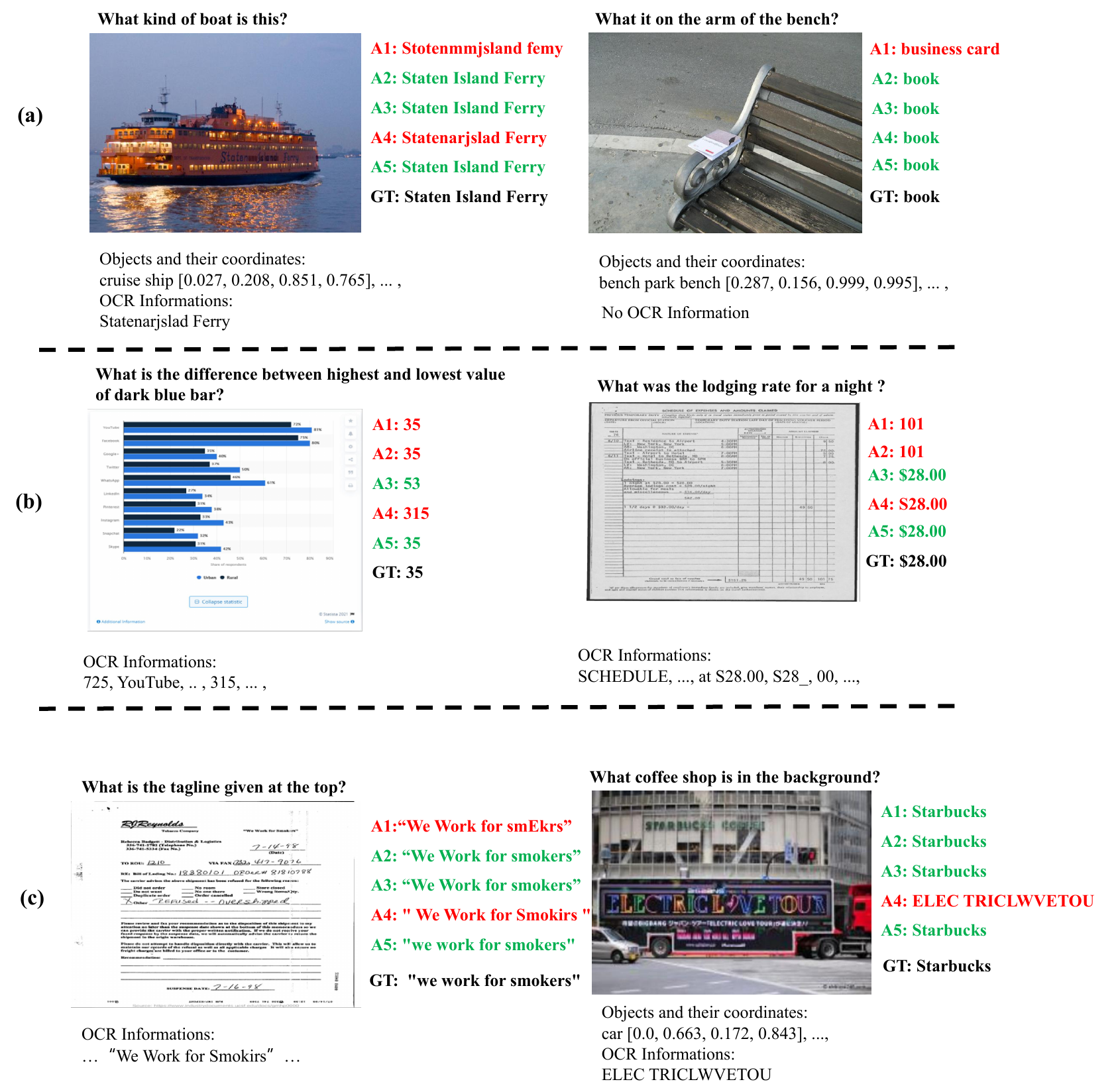}
    \caption{The visualized analysis of proposed modules in IVE. A1 represents the result of using the semantic information encoder only, A2 represents the result of using both the semantic information encoder and low-level information encoder, A3 represents the result of using all three encoders, A4 denotes the result of further integrating the structured knowledge in the inference phrase, A5 denotes the result of integrating the structured knowledge in both the training and inference phrase. GT represents the ground truth. The red lines represent the wrong answers and the green lines denote the correct answers.}
    \label{fig:ab_figure}

\end{figure}

\begin{figure}
    \centering
    \includegraphics[width=1.0\textwidth]{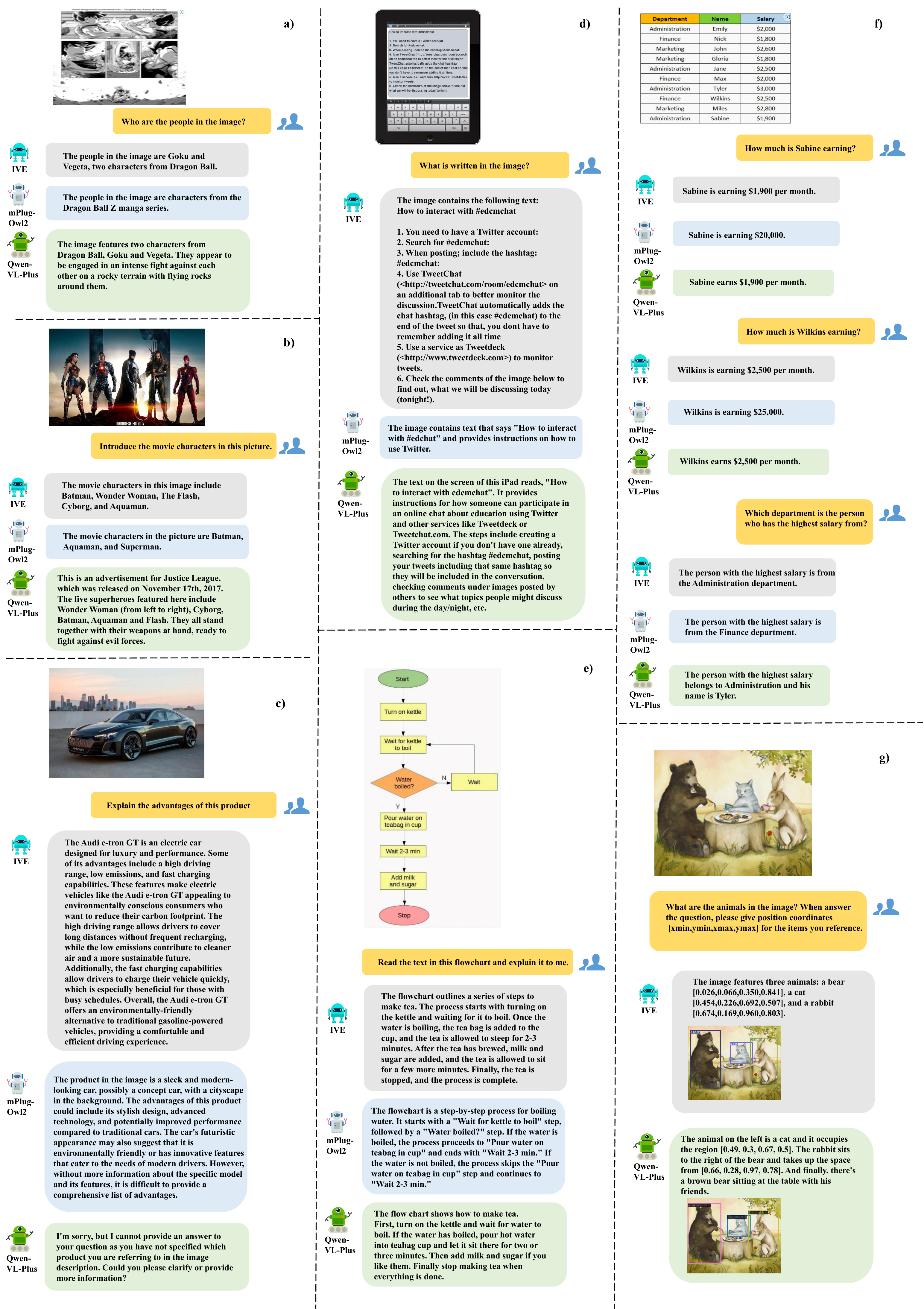}
    \caption{The comparisons among mPLUG-Owl2\cite{ye2023mplug2}, QWen-VL-Plus\cite{bai2023qwen} and our method.}
    \label{fig:gen_image}
\end{figure}
To better evaluate the effects of each proposed module in IVE, we further present the visualized results of the models supervised by different modules in Fig.~\ref{fig:ab_figure}. Among the visualized results, Fig.~\ref{fig:ab_figure}(a) demonstrates that the fusion of the low-level information encoder built upon semantic information encoder is beneficial for the recognition task that requires detailed information. Fig.~\ref{fig:ab_figure}(b) reveals that the further fusion of the document-related information encoder can enhance its understanding of documents and charts. Both Fig.~\ref{fig:ab_figure}(b) and Fig.~\ref{fig:ab_figure}(c) show that the inevitable noises in automatically generated structural knowledge can lead to incorrect responses. However, while integrating the knowledge throughout both the training and inference stages, IVE can resist these noises and generate the correct answers.

Furthermore, we present qualitative results of our model through various examples to showcase the perception capability of our proposed IVE.
Fig.~\ref{fig:gen_image}(a) demonstrates that our method accurately identifies the characters "Goku and Vegeta" in a complex Dragon Ball animation scene, while mPLUG-Owl2\cite{ye2023mplug2} fails to recognize these two characters. Fig.~\ref{fig:gen_image}(b) illustrates IVE can accurately and completely identify five movie characters in the image, whereas mPLUG-Owl2\cite{ye2023mplug2} only identifies three characters and wrongly recognizes a character. 
As shown in  Fig.~\ref{fig:gen_image}(c), IVE generates a richer description compared to mPLUG-Owl2\cite{ye2023mplug2} and QWen-VL-Plus\cite{bai2023qwen}, with the mention of "The Audi e-tron GT" showcasing the advantages of IVE in recogniting details. In Fig.~\ref{fig:gen_image}(d), IVE provides a more complete and accurate description of the textual content on the screen compared to mPLUG-Owl2\cite{ye2023mplug2} and QWen-VL-Plus\cite{bai2023qwen}, reflecting the superior capability of IVE in OCR-related dialogue scenarios. Fig.~\ref{fig:gen_image}(e) involves a flow chart, where IVE accurately describes the relevant steps of "making tea", while the responses generated by mPLUG-Owl2\cite{ye2023mplug2} are confusing. Fig.~\ref{fig:gen_image}(f) demonstrates that IVE can accurately understand the content of a table image, whereas mPLUG-Owl2\cite{ye2023mplug2} cannot. Fig.~\ref{fig:gen_image}(e) and Fig.~\ref{fig:gen_image}(f) illustrate that IVE can successfully comprehend chart images and provide correct answers for each query. Fig.~\ref{fig:gen_image}(g) showcases the ability of IVE in referring grounding tasks, which successfully identifies the categories and corresponding coordinates of three animals in the image.

\end{document}